\documentclass[sn-mathphys,Numbered,iicol,hyperref={pagebackref=true,breaklinks=true,letterpaper=true,colorlinks,bookmarks=false}]{sn-jnl}%

\usepackage{amsthm}%
\usepackage{mathrsfs}%
\usepackage[title]{appendix}%
\usepackage{xcolor}%
\usepackage{textcomp}%
\usepackage{manyfoot}%
\usepackage{booktabs}%
\usepackage{algorithm}%
\usepackage{algorithmic}
\usepackage{listings}%

\usepackage{url}
\usepackage{tikz}
\usepackage{comment}
\usepackage{amsmath,amssymb}
\usepackage{colortbl}
\usepackage{epsfig}
\usepackage[shortlabels, inline]{enumitem}
\usepackage[skip=0pt]{caption}
\usepackage{soul}
\usepackage[utf8]{inputenc}
\usepackage[T1]{fontenc}
\usepackage{amsfonts}
\usepackage{nicefrac}
\usepackage{microtype}

\usepackage{numprint}
\usepackage{siunitx}
\usepackage{etoolbox}
\usepackage{cancel}
\newcommand{\ubold}{\fontseries{b}\selectfont}
\robustify\ubold
\usepackage{bbold}
\usepackage{wrapfig}
\usepackage[skip=2pt]{subcaption}
\usepackage{pifont}
\usepackage{lineno}
\usepackage{array,multirow,graphicx}
\usepackage{indentfirst}
\usepackage{pgfplots}
\pgfplotsset{compat=1.18}

\newcommand{\xmark}{\ding{55}}

\begin{document}

\title[UTOPIA: Unconstrained Tracking Objects without Preliminary Examination via Self-Supervised Domain Adaptation]{\Large UTOPIA: Unconstrained Tracking Objects without Preliminary Examination via Cross-Domain Adaptation}

\author*[1]{\fnm{Pha} \sur{Nguyen}}\email{panguyen@uark.edu}

\author[2]{\fnm{Kha Gia} \sur{Quach}}\email{kquach@ieee.org}

\author[1]{\fnm{John} \sur{Gauch}}\email{jgauch@uark.edu}

\author[3]{\fnm{Samee U.} \sur{Khan}}\email{skhan@ece.msstate.edu}

\author[4]{\fnm{Bhiksha} \sur{Raj}}\email{bhiksha@cs.cmu.edu}

\author*[1]{\fnm{Khoa} \sur{Luu}}\email{khoaluu@uark.edu}

\affil[1]{\orgdiv{Department of CSCE, University of Arkansas, Fayettevile, AR, USA}}

\affil[2]{\orgdiv{pdActive Inc}}

\affil[3]{\orgdiv{Dept. of Electrical and Computer Engineering, Mississippi State University, USA}}

\affil[4]{\orgdiv{Carnegie Mellon University, USA}}

\abstract{Multiple Object Tracking (MOT) aims to find bounding boxes and identities of targeted objects in consecutive video frames. While fully-supervised MOT methods have achieved high accuracy on existing datasets, they cannot generalize well on a newly obtained dataset or a new unseen domain. In this work, we first address the MOT problem from the cross-domain point of view, imitating the process of new data acquisition in practice. Then, a new cross-domain MOT adaptation from existing datasets is proposed without any pre-defined human knowledge in understanding and modeling objects. It can also learn and update itself from the target data feedback. The intensive experiments are designed on four \textit{challenging} settings, including \textit{MOTSynth $\to$ MOT17}, \textit{MOT17 $\to$ MOT20}, \textit{MOT17 $\to$ VisDrone}, and \textit{MOT17 $\to$ DanceTrack}. We then prove the adaptability of the proposed self-supervised learning strategy. The experiments also show superior performance on tracking metrics MOTA and IDF1, compared to fully supervised, unsupervised, and self-supervised state-of-the-art methods.}


\maketitle

\section{Introduction}
\label{sec:intro}

\begin{figure}[t]
    \centering
    \captionsetup[subfigure]{skip=0pt}
    \subcaptionbox{Domain gap visualization.}[0.45\textwidth]{\includegraphics[width=0.45\textwidth]{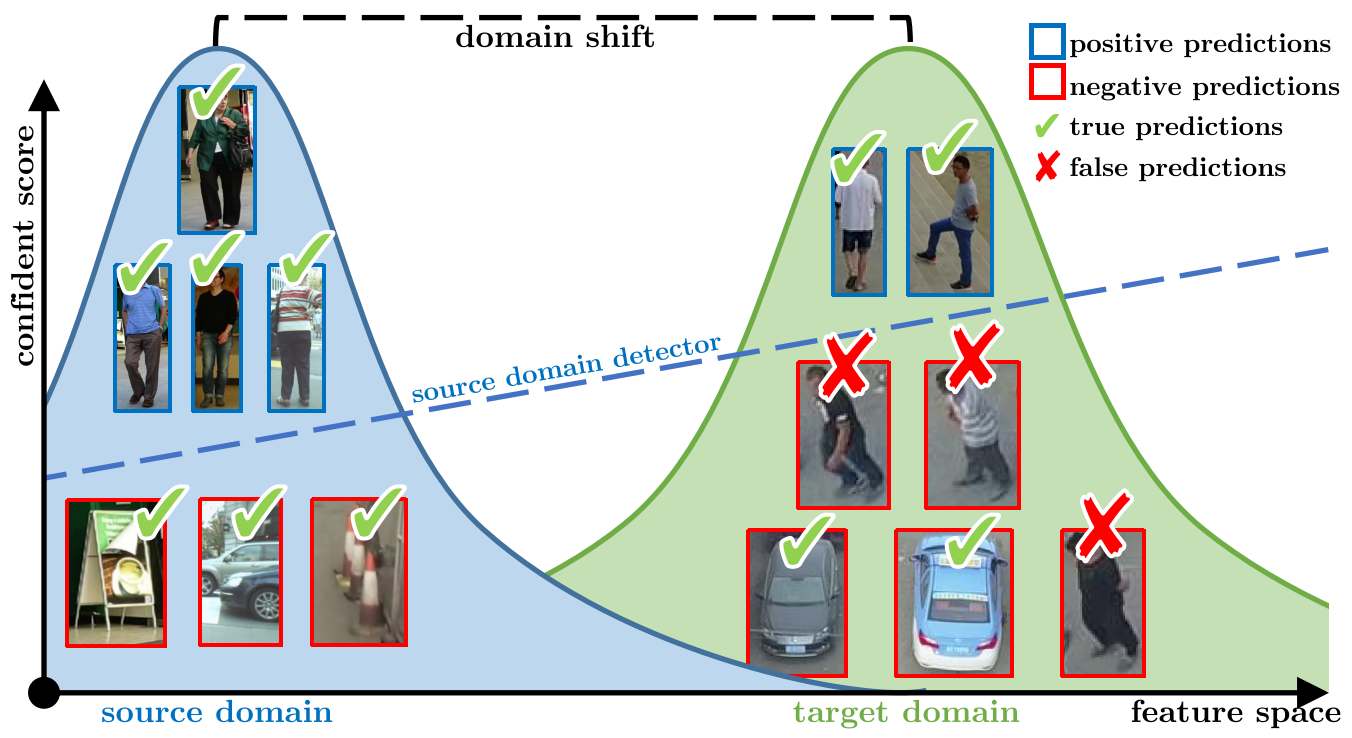}} \hfill
    \captionsetup[subfigure]{skip=0pt}
    \subcaptionbox{\textit{\textit{MOT17 $\to$ VisDrone}}. Ambiguity examples with detected scores.}[0.45\textwidth]{\includegraphics[width=0.45\textwidth]{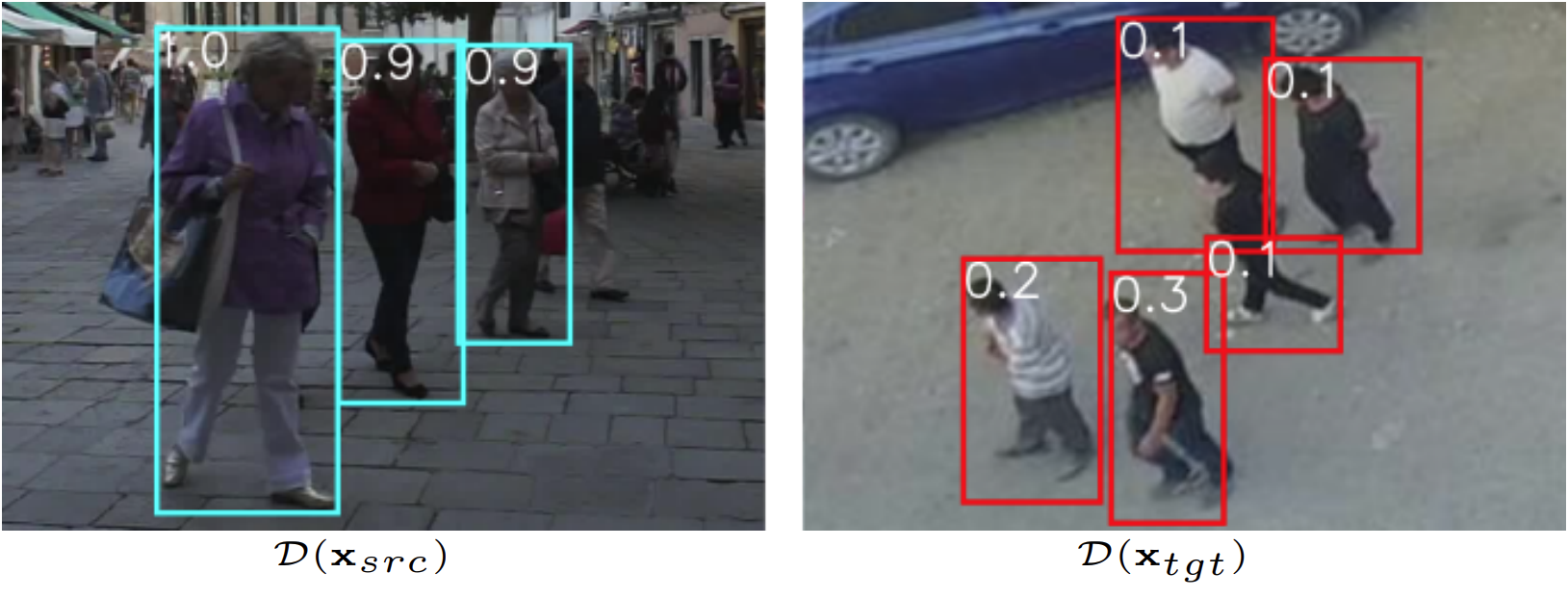}}
    \caption{Directly performing the object detector trained on the source dataset causes ambiguous predictions on the target dataset because of the domain gap. \textbf{Best viewed in color.}}\label{domain_gap}
\end{figure}

Multiple Object Tracking (MOT) has become one of the most critical problems in computer vision. Since 2015, almost every year, a new visual MOT dataset has been introduced~\cite{MOTChallenge2015, MOT16, chen2018bdd100k, caesar2020nuscenes, dendorfer2020mot20, bai2021gmot, sun2022dancetrack, fabbri2021motsynth}. These methods deeply evaluate and inspect numerous challenging aspects of the problem, such as dense view~\cite{dendorfer2020mot20}, identical appearance~\cite{bai2021gmot}, a camera on the move~\cite{chen2018bdd100k, caesar2020nuscenes}, showing its importance and emergence. However, annotating training data for MOT is an intensive and enormously time-consuming task. On average, annotating pedestrian tracks in a six-minute video in the training set of the MOT15~\cite{MOTChallenge2015} requires about 22 hours of manually labeling~\cite{manen2017pathtrack, bastani2021self} using LabelMe tool~\cite{yuen2009labelme}. Given a newly obtained dataset, naively creating pseudo-labels from a model trained on existing datasets is considered an understandable solution~\cite{xiong2021multiview}. However, the performance will be significantly decreased when directly employing an off-the-shell tracker trained on existing datasets, i.e., source domain, to the new dataset, i.e., target domain, without updating feedback signals. It is because of the domain gap in the cross-domain setting, as shown in Fig.~\ref{domain_gap}.

Moreover, MOT is also an extremely non-trivial problem since it requires massive analyses to comprehensively model the given datasets' characteristics. For example, prior works are proposed to study the MOT Challenge~\cite{MOTChallenge2015, MOT16, dendorfer2020mot20, dave2020tao} in detail under a variety of characteristics, including object's type~\cite{rajasegaran2022tracking}, displacement~\cite{zhou2020tracking}, motion~\cite{bewley2016simple, wojke2017simple}, object state~\cite{sun2020simultaneous}, object management~\cite{stadler2021improving}, Bird's-eye-view reconstruction~\cite{dendorfer2022quo}, open-vocabulary~\cite{li2023ovtrack}, and camera motion~\cite{aharon2022bot, Nguyen_2022_CVPR, nguyen2022multi} in long sequences. Nevertheless, the challenge of cross-domain adaptation persists due to a significant domain gap. Therefore, it is necessary to have a ready-to-use method that can adaptively learn and update itself on the target domain without requiring any pre-defined human knowledge in understanding and modeling objects.

Some recent \textit{tracking-by-detection} studies introduced to apply self-supervised learning for training feature extraction models~\cite{bastani2021self, karthik2020simple, yu2022tdt}. However, these methods have not fully solved the self-supervised MOT. By the assumption of having a robust detector, these works opt out of the detection step. Furthermore, they have not intensively explored the cross-domain evaluation setting, a preferable principle and widely used benchmarking in other domain adaptation tasks, i.e., semantic segmentation~\cite{vu2019advent, truong2021bimal}, object detection~\cite{he2022cross}.

\begin{figure}[!t]
    \centering
    \includegraphics[width=1\linewidth]{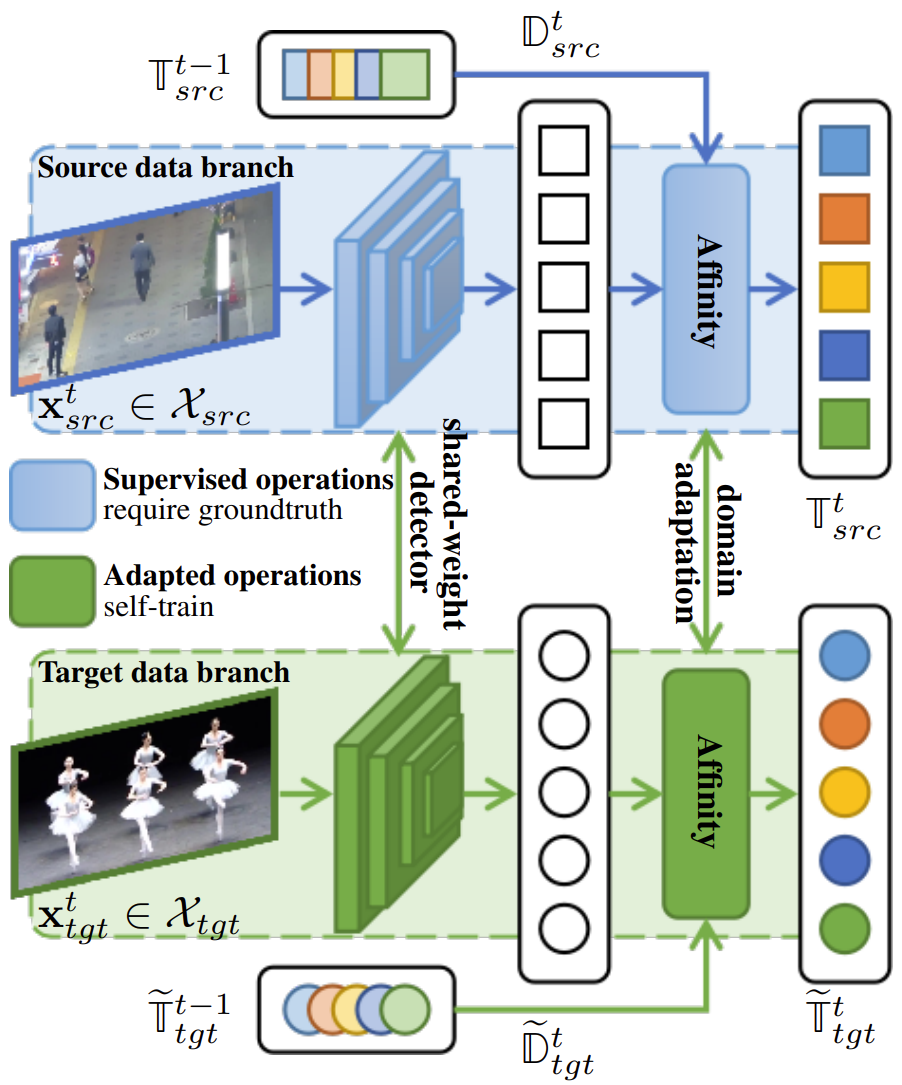}
    \caption{Our proposed UTOPIA to learn self-supervised cross-domain MOT. The proposed method is trained on two data branches simultaneously: source samples (with ground truths) and target samples (without ground truths). The proposed adapted operations will be presented in Section~\ref{sec:method}. Objects presented in circles are samples without ground truth. \textbf{Best viewed in color.}}
    \label{overall}
\end{figure}

To address all these challenges, we propose a novel self-supervised cross-domain learning approach to multiple object tracking, named  \textbf{U}nconstrained \textbf{T}racking \textbf{O}bjects without \textbf{P}reliminary exam\textbf{i}n\textbf{a}tion (UTOPIA). First, a new two-branch deep network attaching both source and target domains will be introduced, as illustrated in Fig.~\ref{overall}. Then a consistency training paradigm will be proposed to guarantee domain discrepancy minimization, leveraging Unsupervised Data Augmentation~\cite{xie2020unsupervised, chen2020simple}. Next, a new proposal assignment mechanism will be presented to learn the similarity. Far apart from prior works~\cite{bastani2021self, karthik2020simple, yu2022tdt}, the proposed entire process is trained end-to-end. In the scope of this paper, we specifically targeted the gap in the camera perspective, synthesized data, occlusion, and appearance. Finally, to prove the substantial generalization of the proposed method, a cross-domain evaluation protocol will be presented to imitate the data acquisition process in practice. To the best of our knowledge, the proposed UTOPIA is one of the first works to introduce MOT in cross-domain conditions. To summarize, the contributions of this work can be listed as follows:
\begin{itemize}[leftmargin=*]
    \item Introduce one of the first studies in cross-domain MOT with the new evaluation settings. Four \textit{challenging} scenarios are chosen so that the target domain poses more challenges than the source domain in many aspects.

    \item Introduce an Object Consistency Agreement (\textbf{OCA}) paradigm to propagate label information from labeled samples to unlabeled ones in the form of a consistency metric and an agreement loss.
    
    \item Present a Optimal Proposal Assignment (\textbf{OPA}) mechanism to self-train the similarity learning. The new Sinkhorn-Knopp Iteration strategy~\cite{cuturi2013sinkhorn} is presented to solve One-to-One and One-to-Many matching, further defined as the objective losses in the tracking deep network.

    \item Achieve substantial improvement in detection and tracking performance compared to numerous methods, including fully supervised, unsupervised, and self-supervised, under the unseen domains.
\end{itemize}

\begin{figure*}[t]
  \centering
  \subcaptionbox{Fully-supervised MOT approach \label{subfig:sublabel1}}[0.45\textwidth]{\includegraphics[width=0.45\textwidth]{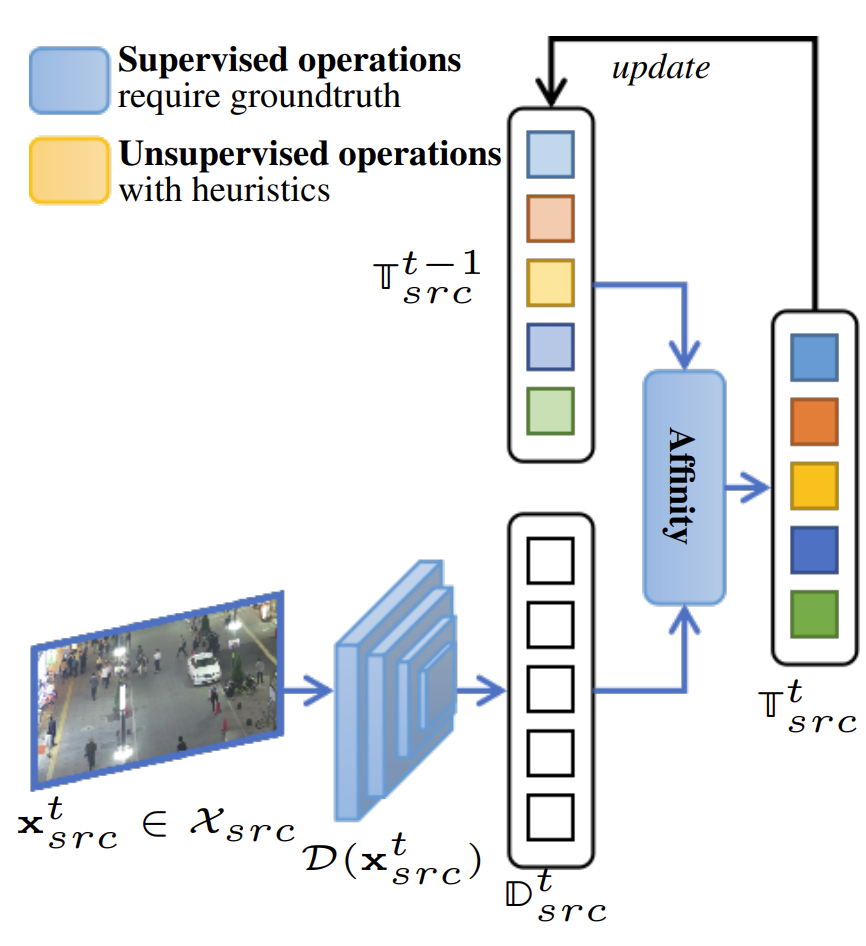}}
  \hfill
  \subcaptionbox{Unsupervised MOT approach \label{subfig:sublabel2}}[0.45\textwidth]{\includegraphics[width=0.45\textwidth]{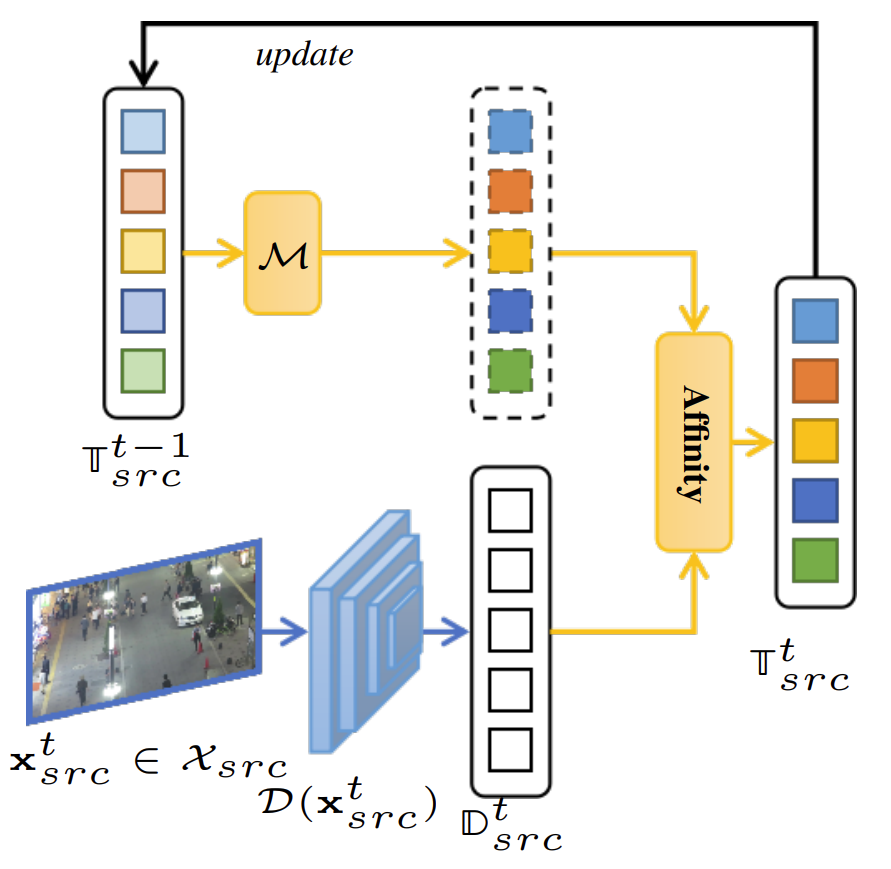}}
  \hfill
\caption{Two common learning types used in most multiple objects tracking methods, including fully-supervised and unsupervised. \textbf{Best viewed in color.}}\label{overview_approaches}
\end{figure*}

In the following sections, we first overview the related works in Section~\ref{sec:related}, then define the problem formulation in Section~\ref{subsec:formulation} and overview the current approaches as illustrated in Fig.~\ref{overview_approaches}. Then a new framework is developed in Section~\ref{sec:method} as illustrated in Fig.~\ref{overall} to simultaneously incorporate the unlabeled data into the entire training process. In Section~\ref{sec:exp}, we conduct experiments to demonstrate the performance of our proposed approach in various domain settings. 

\section{Related Work}
\label{sec:related}

\subsection{Fully-supervised MOT}

\textbf{Learning ID assignment}
Yin et al.~\cite{yin2020unified} trained a Siamese neural network for the joint task of simultaneous single-object tracking and multiple-object association. Rajasegaran et al.~\cite{rajasegaran2022tracking} lifted people's 3D information to represent the 3D pose of the person, their location in the 3D space, and the 3D appearance then computed the similarity between predicted states and observations in a probabilistic manner.

\textbf{Learning object's motion} Xiao et al.~\cite{xiao2018simple} adopted an optical flow network to estimate the object's location. Zhou et al.~\cite{zhou2020tracking} employed a straightforward approach, which trained a network to predict the movement offset from the previous frame and then matched it with the nearest tracklet center point. Bergmann et al.~\cite{bergmann2019tracking} showed a simple approach by exploiting the bounding box regression of the object detector to guess the position of objects in the next frame in a high-frame-rate video sequence without camera motion. Sun et al.~\cite{sun2020simultaneous} constructed three networks to compute three matrices representing the object's motion, type, and visibility for every matching step.

\textbf{Joint detection and tracking}
Chan et al.~\cite{CHAN2022108793} proposed an end-to-end network for simultaneously detecting and tracking multiple objects.
Pang et al.~\cite{pang2021quasi} presented a combination of similarity learning and other detection methods~\cite{girshick2015fast, ren2015faster}, which densely samples many region proposals on a single pair of images. Meinhardt et al.~\cite{meinhardt2022trackformer} introduced a new tracking-by-attention mechanism with data association via attention between the frames.
Wu et al.~\cite{wu2021track} presented a joint online detection and tracking model which explores tracking information during inference to guide the detection and segmentation.
Yan et al.~\cite{yan2022towards} presented a unified model to solve four tracking problems with a single network and the same parameters. It maintains the same input, backbone, head, and embedding among all tracking tasks.

\subsection{Unsupervised MOT by Heuristics}

\textbf{Heuristics on ID assignment} Zhang et al.~\cite{zhang2021bytetrack} introduced a generic tracking method to associate all the detection boxes, including low-confident bounding boxes, instead of only the high-scored boxes. In the case of low-score boxes, they use similar tracklets to recover proper objects. Stadler et al.~\cite{stadler2021improving} proposed a novel occlusion handling strategy that explicitly models the relation between occluding and occluded tracks in both temporal directions while not depending on a separate re-identification network.

\textbf{Assumptions on object's motion} Kalman filter is one of the most used methods to model linear object's velocity (i.e., in~\cite{bewley2016simple, wojke2017simple, cao2022observation, aharon2022bot}).
Cao et al.~\cite{cao2022observation} showed that a simple motion model could better track without the appearance information. They emphasized observation during the loss of recovery tracks to reduce the error.
Aharon et al.~\cite{aharon2022bot} proposed a camera motion compensation-based features tracker and a suitable Kalman filter state vector for better box localization.

\subsection{Self-supervised MOT}
Many works assume having a robust object detector and only focus on training a self-supervised feature extractor. Bastani et al.~\cite{bastani2021self} proposed a method to train a model to produce consistent tracks between two distinct inputs from the same video sequence. Karthik et al.~\cite{karthik2020simple} presented a method to generate tracking labels using SORT~\cite{bewley2016simple} for given unlabeled videos. They used a ReID network with Cross-Entropy loss to predict the generated labels. Yu et al.~\cite{yu2022tdt} combined both one-stage and two-stage methods to predict the detections and their embeddings with the help of distillation. Valverde et al.~\cite{valverde2021there} presented a framework consisting of multiple teacher networks, each of which takes a specific modality as input, i.e., RGB, depth, and thermal, to maximize the complementary cues, i.e., appearance, geometry, and reflectance.

\textbf{Difference from Previous Works} Prior works remove the error-prone detection step using ground-truth bounding boxes for the self-supervised setting and focus on the self-supervised contrastive learning for the tracklet association step. We instead propose an end-to-end self-learning framework, from object detection to similarity learning, and show its substantial generalization by performing on four \textit{challenging} data settings presented in the sections below.

\section{Preliminaries}
\label{subsec:formulation}

\subsection{Problem Definition}

We denote $\mathbf{x}^t_{src}$ is a source sample at a particular time step $t$ in the source domain scenes $\mathcal{X}_{src}$, here $\mathbf{x}^t_{src} \in \mathcal{X}_{src}$ and $\mathcal{X}_{src} \subset \mathbb{R}^{W\times H\times 3}$ the image space.
Along with each sample $\mathbf{x}^t_{src}$, a set of ground-truth objects $\mathbb{O}_{src}^{t} = \{ \mathbf{o}^{t}_{i}\}$ associated with their locations and identities. 
The ground-truth object is denoted as $\mathbf{o}^{t}_{i} = (o_x, o_y, o_w, o_h, o_{id})$. 
Let $\mathcal{D}$ be the object detector, which takes an input sample $\mathbf{x}^t_{src}$ and produces a list of detections $\mathcal{D}(\mathbf{x}^t_{src}) = \mathbb{D}_{src}^{t} = \{ \mathbf{d}^{t}_{j} \;|\; 0 \leq j < M \}$ by localizing and estimating the proposal regions to obtain locations, sizes and foreground confident scores $\mathbf{d}^{t}_{j} = (d_x, d_y, d_w, d_h, d_{score})$, thresholding $d_{score} \geq \gamma$. To determine the identity of each proposal $\mathbf{d}^{t}_{j}$, we denote $\mathcal{T}$ as the multiple object tracker, and $\mathbb{T}_{src}^{t}$ as the set of tracklets at the time step $t$, which contains detected objects with consistent identity throughout the period. We define: $\mathbb{T}_{src}^{t} = \{ \mathbf{tr}^{t}_{k} = (tr_x, tr_y, tr_w, tr_h, tr_{id}) \;|\; 0 \leq k < N \}$,
The object tracker takes the previous object states and the currently detected objects and then performs an affinity step to update new states as in Eqn.~\eqref{eq:newstate}.
\begin{equation} \label{eq:newstate}
  \mathbb{T}_{src}^{t} =
  \begin{cases}
    \texttt{initialize}(\mathbb{D}_{src}^{t})                 & \text{if } t = 0 \\
    \mathcal{T}(\mathbb{T}_{src}^{t-1}, \mathbb{D}_{src}^{t}) & \text{if } t > 0
  \end{cases}
\end{equation}
In general, there are many approaches proposed to solve the equation $\mathbb{T}_{src}^{t} = \mathcal{T}(\mathbb{T}_{src}^{t-1}, \mathbb{D}_{src}^{t})$ in different ways~\cite{meinhardt2022trackformer, pang2021quasi, zhang2021bytetrack, zhou2020tracking, weng2022whose}.
Without loss of generality, these approaches can be divided into two categories, i.e., fully supervised and unsupervised methods. In fully-supervised approaches, Fig.~\ref{subfig:sublabel1} illustrates the processing flow, and the equation is formulated as in Eqn.~\eqref{fullysupervised}.
\begin{equation}
  \mathbb{T}_{src}^{t} = \texttt{argmax}\Bigg(\texttt{sim}\Big(\mathcal{F}(\mathbb{T}_{src}^{t-1}), \mathcal{F}(\mathbb{D}_{src}^{t})\Big)\Bigg)\label{fullysupervised}
\end{equation}
where $\mathcal{F}$ is a feature extractor, which can simply be an RoI pooling layer as in~\cite{girshick2015fast, ren2015faster} or a Re-Identification model~\cite{he2020fastreid, wang2021different}. In other words, these approaches learn a similarity function \texttt{sim} to calculate the probability of merging a detection and a tracklet based on their deep features.

On the other hand, the unsupervised approach is shown in Fig.~\ref{subfig:sublabel2}, and it formulates the solution as in Eqn.~\eqref{eq:tracklet}.
\begin{equation}
  \mathbb{T}_{src}^{t} = \texttt{argmax}\Bigg(\texttt{IoU}\Big(\mathcal{M}(\mathbb{T}_{src}^{t-1}), \mathbb{D}_{src}^{t}\Big)\Bigg)
  \label{eq:tracklet}
\end{equation}
where $\mathcal{M}$ is a non-parametric motion model estimating an object's future state based on previous states, i.e., Kalman Filter, as used in~\cite{bewley2016simple, wojke2017simple, rajasegaran2022tracking, stadler2021improving}.

Besides, there are also some fully-supervised variants using a parametric motion model $\underset{\theta}{\mathcal{M}}$ (i.e., visual offset~\cite{zhou2020tracking}, LSTM~\cite{chaabane2021deft}, attention~\cite{weng2022whose}, transformer~\cite{Nguyen_2022_CVPR}), and supervised-unsupervised crossovers~\cite{wojke2017simple, zhang2021bytetrack}.

\subsection{Limitations of Fully-supervised Losses}

\textbf{Object Detection} Given ground-truth objects $\mathbb{O}_{src}^{t}$, the Smooth $\ell_1$ distance and Cross-Entropy loss are adopted to effortlessly learn two supervised tasks, i.e., bounding box regression and object classification, respectively as in Eqn.~\eqref{location_loss}.
\begin{equation}
  \small
  \begin{split}
    \mathcal{L}_{det} = \frac{1}{|\mathbb{D}_{src}|}\sum_{\mathbf{d}_{i} }^{\mathbb{D}_{src}}\Big[\lambda_{reg}\ell_1(\mathbf{o}^{+}, \mathbf{d}_{i}) + \lambda_{cls}\ell_{CE}(\mathbf{o}^{+}, \mathbf{d}_{i})\Big]
  \end{split}
  \label{location_loss}
\end{equation}
where $\lambda_{reg}$, $\lambda_{cls}$ are weighted parameters to balance corresponding objective functions, $\mathbf{d}_{i}$ is a object proposal, $\mathbf{o}^{+}$ is the \textit{positive} ground-truth object that have maximum IoU with that proposal $\mathbf{d}^{t}_{i}$, following~\cite{girshick2015fast, ren2015faster}.

\textbf{Similarity Learning} 
In order to train the instance similarity, some approaches use an off-the-shelf Re-Identification model~\cite{bergmann2019tracking, wojke2017simple, zhang2021bytetrack}.
The Softmax with Cross-Entropy loss function~\cite{wu2018unsupervised, oord2018representation} to train the feature extractor $\mathcal{F}$ is then defined as in Eqn.~\eqref{sup_sim}.
\begin{equation}
  \small
  \begin{split}
    \mathcal{L}_{sim} = \frac{1}{|\mathbb{T}_{src}|}\sum_{\mathbf{tr}_{i} }^{\mathbb{T}_{src}}\log\Bigg\{1+\sum_{\mathbf{o}^{-}}\Big[\exp\Big(\mathcal{F}(\mathbf{tr}_{i})\cdot\mathcal{F}(\mathbf{o}^{-})\Big) \\ -\exp\Big(\mathcal{F}(\mathbf{tr}_{i})\cdot\mathcal{F}(\mathbf{o}^{+})\Big)\Big]\Bigg\}
  \end{split}
  \label{sup_sim}
\end{equation}
where $\mathbf{tr}_{i}$ is drawn from the tracker's output set $\mathbb{T}_{src}$ as an anchor, and $\mathbf{o}^{-}$ are \textit{negative} ground-truth objects drawn from $\mathbb{O}_{src}$. These \textit{negative} objects are all remaining objects other than $\mathbf{o}^{+}$. 

However, in the self-supervised setting, the components $\mathbf{o}^{+}$ and $\mathbf{o}^{-}$ in Eqn.~\eqref{location_loss} and Eqn.~\eqref{sup_sim} are missing, so the losses could not be calculated.
A new strategy for making full use of the ambiguity or uncertainty predictions~\cite{he2019bounding, zhang2021bytetrack, pang2021quasi} and enhancing the certainty in selecting those missing components will be introduced to address the incalculability problem in the Subsection~\ref{subsec:proposal}. Furthermore, although the Eqn.~\eqref{sup_sim} is a fundamental loss that is widely used, it elevates the unbalance in the number of positive and negative samples. Only one positive sample can be matched, while multiple negative samples are considered. This problem can be solved in our One-to-Many matching strategy via Optimal Transport and Multiple-Positive loss presented in our Subsection~\ref{subsec:assignment}.

\subsection{Optimal Transport in ID Assignment}\label{OT_IDassignment}

After obtaining a good similarity representation model guided by the Eqn.~\eqref{sup_sim}, the next step is to assign the object identity. We use the Optimal Transport method to develop our ID Assignment strategy. While the same objective methods, i.e., the Hungarian algorithm, can only estimate hard-matching pairs in a fixed One-to-One assignment manner, we instead explore the usability of Optimal Transport in both One-to-One and One-to-Many strategies that are well fit in our problem. 
Let $\mathbf{C} (\mathbb{T}^{t-1}_{src}, \mathbb{D}^t_{src}) = (c[{i, j}])$ be the transportation cost matrix where $c[{i, j}]$ measures the cosine distance to associate from $\mathbf{tr}^{t - 1}_{i}$ to $\mathbf{d}^{t}_{j}$ as in Eqn.~\eqref{eq:disassoc}.
\begin{equation}
    c[{i, j}] = 1 - \frac{\mathcal{F}(\mathbf{tr}^{t - 1}_{i})^\intercal\cdot\mathcal{F}(\mathbf{d}^{t}_{j})}{||\mathcal{F}(\mathbf{tr}^{t - 1}_{i})|| \; || \mathcal{F}(\mathbf{d}^{t}_{j})||}
    \label{eq:disassoc}
\end{equation}
where $i$ and $j$ are the indexers for the rows and columns, which will be used in the rest of the paper.

Optimal Transport addresses the problem of finding the best assignment solution $\pi$ in the set of all possible couplings $\Pi (\mathbf{p}, \mathbf{q}) = \{ \pi \in \mathbb{R}^{N \times M} \; | \; \pi \mathbb{1}^M =\mathbf{p}, \pi^\intercal \mathbb{1}^N =\mathbf{q} \}$ to transport the mass that minimizes the transportation cost between two distributions as in Eqn.~\eqref{eq:optimal_transport}.
\begin{equation}
  \label{eq:optimal_transport}
  \min_{\pi \in \Pi(\mathbf{p}, \mathbf{q})} \sum_{i}^N \sum_{j}^M c[{i, j}]\pi[{i, j}]
\end{equation}
where $\mathbf{p}$ and $\mathbf{q}$ are the marginal weights, which are attached to $\pi$ on its rows and columns, respectively. From the formulation for Optimal Transport-based Assignment, as defined in Eqn.~\eqref{eq:optimal_transport}, it can be solved as a linear programming problem.

Optimal Transport is a well-studied topic in Optimization Theory and recently received attention in Computer Vision due to its potential in many relevant topics, i.e., visual matching~\cite{sarlin2020superglue}, object detection~\cite{ge2021ota}, in-flow and out-flow counting~\cite{han2022dr}. The method explores not only the continuity and the differentiability~\cite{fragala2005continuity, di2020optimal} but also the flexibly optimal assigning strategy~\cite{dong2020copt, 9762998} in an end-to-end training network.

However, when there are multiple proposals and sampling bounding boxes, i.e., $N \times 1000$ in our One-to-Many setting, the resulting linear program can be very low-efficient by the polynomial time complexity. This problem will be addressed in the Subsection~\ref{subsec:iteration_algo}.

\begin{figure*}[!t]
    \centering
    \includegraphics[width=1.0\linewidth]{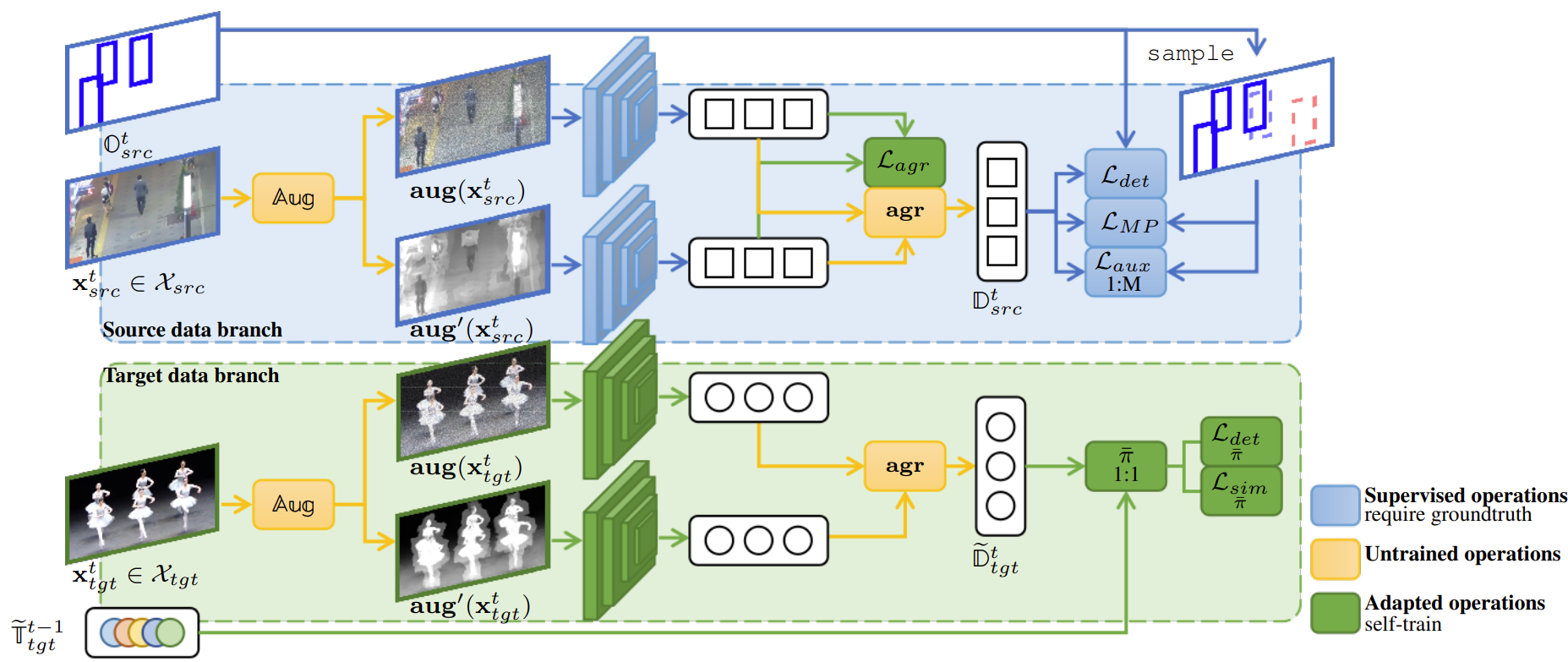}
    \caption{Our proposed UTOPIA training flow consists of two data branches trained simultaneously. $\underset{\bar{\pi}}{\mathcal{L}_{det}}$ and $\underset{\bar{\pi}}{\mathcal{L}_{sim}}$ are computed based on the selection strategy $\underset{1:1}{\bar{\pi}}$, so we consider them as adapted operations. \textbf{Best viewed in color.}}
    \label{overall_framework}
\end{figure*}

\subsection{Unsupervised Data Augmentation}

Given a new sample from an unseen domain different from the training source domain, a trained object detector is usually unable to produce a high-confident prediction
as illustrated in Fig.~\ref{domain_gap}. However, as a result of taking low-confident objects into account, the false positive rate also increases. To mitigate the trade-off between sensitivity and specificity, Unsupervised Data Augmentation (UDA)~\cite{xie2020unsupervised, chen2020simple} is inspected in teaching the detector to consistently recognize objects over many data augmentation methods applied in source samples $\mathbf{x}^t_{src}$, furthermore enhance the precision rate in detecting objects from target samples $\mathbf{x}^t_{tgt}$.

UDA presents a mechanism to propagate label information from labeled to unlabeled examples. It originally injects noise or a simple augmentation $\mathbf{aug}(\cdot)$ into an unlabeled sample $\mathbf{x}_{tgt}$. Then it optimizes the consistency objective between them via Cross-Entropy loss as in Eqn.~\eqref{eq:ce_loss}.
\begin{equation}
    \mathcal{L}_{UDA} = \ell_{CE}\Bigg(\mathcal{F}(\mathbf{x}_{tgt}), \mathcal{F}\Big(\mathbf{aug}(\mathbf{x}_{tgt})\Big)\Bigg)
    \label{eq:ce_loss}
\end{equation}

Although the loss function in Eqn.~\eqref{eq:ce_loss} influences the consistency in the feature space, it cannot regulate the detection problem. Inspired by UDA, a new agreement loss is introduced for complex scenes containing multiple objects as in Subsection~\ref{subsec:proposal}.

\section{The Proposed Approach}
\label{sec:method}

On the target domain $\mathcal{X}_{tgt}$, we propose the new Object Consistency Agreement (\textbf{OCA}) approach, as in Subsection~\ref{subsec:proposal}, to maximize the consistency of the object's existence, and the new Optimal Proposal Assignment (\textbf{OPA}), as in Subsection~\ref{subsec:assignment}, to adaptively train the similarity learning process.
The proposed training flow is shown in Fig.~\ref{overall_framework}. %

\subsection{Object Consistency Agreement (OCA)}
\label{subsec:proposal}

Randomly drawing two augmentation methods $\mathbf{aug}$ and $\mathbf{aug}^\prime$ from augmentation set $\mathbb{Aug}$ and applying to an input image $\mathbf{x}^t_{src}$. Initially, the detection loss in Eqn.~\eqref{location_loss} has to be held and optimized, i.e., $\mathcal{L}_{det}\Bigg(\mathcal{D}\Big(\mathbf{aug}(\mathbf{x}^t_{src})\Big)\Bigg) + \mathcal{L}_{det}\Bigg(\mathcal{D}\Big(\mathbf{aug}^\prime(\mathbf{x}^t_{src})\Big)\Bigg)$. %

The agreement metric is defined for differently augmented views of the same data sample as a GIoU~\cite{Rezatofighi_2018_CVPR} cost matrix as in Eqn.~\eqref{eq:costmatrix}.
\begin{equation}
    \mathbf{agr}(\mathbf{x}^t_{src}) = \texttt{GIoU}\Bigg(\mathcal{D}\Big(\mathbf{aug}(\mathbf{x}^t_{src})\Big), \mathcal{D}\Big(\mathbf{aug}^\prime(\mathbf{x}^t_{src})\Big)\Bigg)
    \label{eq:costmatrix}
\end{equation}
and take that agreement metric as a loss function:
\begin{equation}
    \mathcal{L}_{agr} = \underset{i}{\texttt{avg}}\Bigg(1 - \underset{j}{\texttt{max}}\Big(\mathbf{agr}(\mathbf{x}^t_{src})\Big)\Bigg)
    \label{eq:lossfunc}
\end{equation}
In other words, two separate stochastic transformations, which are applied to any given data sample, first smoothen the model's prediction with respect to changes in the Input. With a good selection of augmentation methods $\mathbb{Aug}$, the model successfully produces consistent prediction over two stochastic transformations meaning that it is one step closer to bridging the domain gap between source and target. The agreement loss $\mathcal{L}_{agr}$ is added to guarantee this learning process. In this selection, we present our investigation and recommendation in the ablation study section~\ref{subsec:ablation}. Keeping original input image $\mathbf{aug}(\mathbf{x}^t_{src}) = \mathbf{x}^t_{src}$, termed \textit{identity} operation, is by default included in the $\mathbb{Aug}$ set.

\begin{figure*}[t]
  \centering
  \subcaptionbox{The input \texttt{GIoU} cost (lower is better)}[0.3\textwidth]{\includegraphics[width=0.3\textwidth]{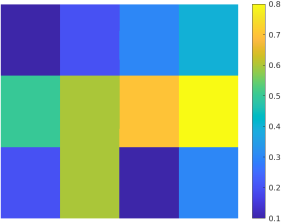}}
  \hfill
  \subcaptionbox{The optimized one-to-one plan $\bar{\pi}$ (higher is better)}[0.3\textwidth]{\includegraphics[width=0.3\textwidth]{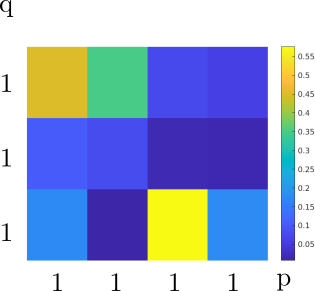}}
  \hfill
  \subcaptionbox{The optimized one-to-many plan $\bar{\pi}$ (higher is better)}[0.3\textwidth]{\includegraphics[width=0.3\textwidth]{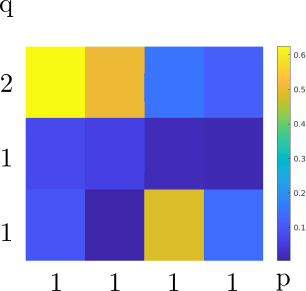}}
  \hfill
\caption{Input and outputs for each optimization strategy of the Sinkhorn-Knopp Iteration algorithm \cite{cuturi2013sinkhorn} in our implementation. \textbf{Best viewed in color.}}\label{optimize}
\end{figure*}

Alternatively, the agreement is employed as a metric in the proposal selection strategy on the target domain. Let $\widetilde{\mathbb{D}}_{tgt}$ be the list of detections which is an extended set of $\mathbb{D}_{tgt}$, additionally containing low confident detections $\widetilde{\mathbb{D}}_{tgt} = \{(\widetilde{d}_x, \widetilde{d}_y, \widetilde{d}_w, \widetilde{d}_h, \widetilde{d}_{score}) \; | \; \widetilde{d}_{score} \geq \widetilde{\gamma}\}$. Here $\gamma > \widetilde{\gamma}$ is a low threshold (i.e. $0.1$):
\begin{equation}
   \widetilde{\mathbb{D}}_{tgt} = \underset{j}{\texttt{argmax}}\Big(\mathbf{agr}(\mathbf{x}^t_{tgt})\Big)\label{agr_metric}
\end{equation}
\texttt{argmax} returns a list of indices used to obtain $\widetilde{\mathbb{D}}_{tgt}$ via indexing. Object features are extracted on the original image $\mathcal{F}(\mathbf{x}^t_{tgt})$ and then the tracker $\mathcal{T}$ is performed: $\widetilde{\mathbb{T}}^t_{tgt} = \mathcal{T}(\widetilde{\mathbb{T}}_{tgt}^{t-1}, \widetilde{\mathbb{D}}_{tgt}^{t})$.

The agreement metric in Eqn.~\eqref{agr_metric} calculated on object proposals in a new domain, even with the low confident $\widetilde{\mathbf{d}}$ ones, indicates the existence of objects. In this step, we empirically pick maximum-intersection pairs with GIoU to score $\texttt{GIoU}[i, j]$ more significant than $0.4$, then perform non-maximum suppression to get the final bounding boxes.

\subsection{Sinkhorn-Knopp Iteration Algorithm}\label{subsec:iteration_algo}

The polynomial time complexity in the Subsection~\ref{OT_IDassignment} can be addressed by a fast iterative solution named Sinkhorn-Knopp~\cite{cuturi2013sinkhorn}. It converts the optimization target in Eqn.~\eqref{eq:optimal_transport} into a non-linear but convex form using a regularization term $E$ as in Eqn.~\eqref{eq:ot_reg}.
\begin{equation}
  \label{eq:ot_reg}
  \min_{\pi \in \Pi(\mathbf{p}, \mathbf{q})} \sum_{i}^N \sum_{j}^M c[{i, j}] \pi[{i, j}] + \gamma E\left( \pi[{i, j}] \right)
\end{equation}
where $E( \pi[{i, j}] ) = \pi[{i, j}] ( \log(\pi[{i, j}]) - 1)$, and $\gamma$ is a learnable parameter, initially set to 0.5 and used to control the intensity of the regulation. The iteration algorithm in Eqn.~\eqref{eq:iteration} as implemented in~\cite{han2022dr, sarlin2020superglue} updates the cost.
\begin{equation} \label{eq:iteration}
u^{t+1}_j = \frac{q_j}{\sum_i W_{ij} v^t_i} \;, \; v^{t+1}_i = \frac{p_i}{\sum_j W_{ij} u^t_j}
\end{equation}
where $v$ and $u$ are two non-negative vectors of scaling coefficients~\cite{ge2021ota}.

After repeating this iteration multiple times, i.e., $100$ in our experiments, the approximate optimal plan $\bar{\pi}$ can be obtained as in Eqn.~\eqref{eq:sinkhorn_iteration_final}.
\begin{equation} \label{eq:sinkhorn_iteration_final}
  \bar{\pi} = \text{diag}(v) \mathbf{W} \text{ diag}(u)
\end{equation}
where $\mathbf{W} = e^{-\frac{1}{\gamma}\mathbf{C}}$. The higher the returned value $\bar{\pi}[{i, j}]$, the more units are recommended to be transported. In other words, the more likely that two samples should be matched. 
We provide a matching sample to intuitively illustrate Input and outputs for each optimization strategy of the Sinkhorn-Knopp Iteration algorithm \cite{cuturi2013sinkhorn} in our implementation as shown in Fig. \ref{optimize}. The marginal weights (i.e., $\mathbf{p}$ and $\mathbf{q}$) controlling the total supplying units are attached to the sides of the matrices.

\subsection{Optimal Proposal Assignment (OPA)}
\label{subsec:assignment}

\textbf{One-to-One (1:1) Assignment} The marginal weights (i.e. $\mathbf{p}$ and $\mathbf{q}$) control the total supplying units:
\begin{equation}
    \mathbf{p}[i] = \sum^M_j\pi[i, j] \text{ and } \mathbf{q}[j] = \sum^N_i\pi[i, j]
\end{equation}

On the target domain, when matching two output $\widetilde{\mathbb{T}}^{t-1}_{tgt}$ and $\widetilde{\mathbb{D}}^t_{tgt}$ of two consecutive frames, one sample should be associated with another sample, so $\mathbf{p} = \mathbb{1}^N$ and $\mathbf{q} = \mathbb{1}^M$. We use $\underset{\bar{\pi}}{\mathcal{L}_{det}}$ as in Eqn.~\eqref{location_loss} and $\underset{\bar{\pi}}{\mathcal{L}_{sim}}$ as in Eqn.~\eqref{sup_sim} to train the network, positive and negative soft-labels are balanced by choosing one sample for each type, and selected based on the optimal plan $\bar{\pi}$, where $\mathbf{o}^{+}$ and $\mathbf{o}^{-}$ now are replaced by $\underset{j}{\texttt{argmax}}(\bar{\pi})$ and $\underset{j}{\texttt{argmin}}(\bar{\pi})$.

\textbf{One-to-Many (1:M) Assignment}
On the source domain $\mathcal{X}_{src}$ where ground-truth boxes are provided, a proposal sampler can be used to firstly guarantee the balanced number of positive and negative bounding boxes, secondly, provide more informative observations to the network for similarity learning. We adapt the cost to $\mathbf{C} (\mathbb{T}^t_{src}, \texttt{sample}(\mathbb{O}^{t}_{src}))$ by using the IoU sampler~\cite{chen2019mmdetection}. For that \texttt{sample} operation, we know the number of positive samples $\texttt{sample}^{+}(\mathbb{O}_{src})$ and negative samples $\texttt{sample}^{-}(\mathbb{O}_{src})$, so the values of $\mathbf{p}$ and $\mathbf{q}$ now become:
\begin{equation}
\begin{gathered}
    \mathbf{p}[i] = |\texttt{sample}^{+}(\mathbf{o}^{+}) | \quad, \\ 
    \mathbf{q}[j] = 
    \begin{cases}
    1 \text{ if } \mathbf{o}_{j} \in \texttt{sample}^{+}(\mathbb{O}_{src}) \\
    0 \text{ if } \mathbf{o}_{j} \in \texttt{sample}^{-}(\mathbb{O}_{src})
    \end{cases}
\end{gathered}
\end{equation}

The Multiple-Positive loss function~\cite{sun2020circle, pang2021quasi} is then adapted from Eqn.~\eqref{sup_sim} to train this scenario:
\begin{equation}
  \small
  \begin{split}
    \mathcal{L}_{MP} = \log\Bigg\{1+\sum_{\mathbf{o}^{+} }\sum_{\mathbf{o}^{-}}\Big[\exp\Big(\mathcal{F}(\mathbf{tr}_{i})\cdot\mathcal{F}(\mathbf{o}^{-})\Big) \\ -\exp\Big(\mathcal{F}(\mathbf{tr}_{i})\cdot\mathcal{F}(\mathbf{o}^{+})\Big)\Big]\Bigg\}
  \end{split}
\end{equation}

In this branch, optimal plan $\bar{\pi}$ is used as an auxiliary loss in addition to the Multiple-Positive loss function with ground-truth matches:
\begin{equation}
  \small
    \underset{\text{\scriptsize 1:M}}{\mathcal{L}_{aux}} = \bar{\pi}[i, j] - c \text{ where } c = 
    \begin{cases}
    1\text{ if }\mathbf{o}_{j} \in \texttt{sample}^{+}(\mathbb{O}_{src}) \\
    0\text{ if }\mathbf{o}_{j} \in \texttt{sample}^{-}(\mathbb{O}_{src})
    \end{cases}
\end{equation}

\section{Experimental Results}
\label{sec:exp}

\subsection{Datasets}
\label{subsec:datasets}

\textbf{MOT Challenge}~\cite{dendorfer2020mot20, MOT16} is a commonly used benchmarking dataset for pedestrian tracking. This dataset has two versions, including MOT17~\cite{MOT16} and MOT20~\cite{dendorfer2020mot20}. Each set consists of real-world surveillance and handheld camera footage with various challenging conditions, such as occlusions, crowded walking people, viewing angles, illuminations, and frame rates. 

\textbf{MOTSynth}~\cite{fabbri2021motsynth} is a large-scale synthetic dataset comprising 768 video sequences for detection, tracking, and segmentation problems. Each video sequence is generated by the GTA-V game with various pedestrian models in different clothes, backpacks, bags, masks, hair, and beard styles. Each frame contains 29.5 people on average and 125 people at max, with over 9,519 unique pedestrian identities.

\textbf{VisDrone}~\cite{visdrone} contains 288 video sequences captured by cameras mounted on various types of drones. The dataset was collected in different scenarios and under various weather and lighting conditions. There are more than 2.6 million manually annotated bounding boxes of objects of interest, including pedestrians, cars, bicycles, and tricycles.

\textbf{DanceTrack}~\cite{sun2022dancetrack} contains 100 dance videos of different dance genres, including classical dance, street dance, pop dance, large group dance, and sports. This dataset is more challenging for motion-based tracking approaches since the object motion is highly non-linear frequently occluding and crossing over each other.

\subsection{Experimental Setups}
\label{subsec:setup}

To demonstrate the robustness of UTOPIA, we construct four \textit{challenging} cross-domain scenarios on MOT datasets described in~\ref{subsec:datasets}.

\textbf{Scenario 1 -- from synthesized to real-data:} MOTSynth~\cite{fabbri2021motsynth} is used as the source train. The target train is MOT17 \textit{half-train} while MOT17 \textit{half-val} is used as a validation set with \textit{half-train} and \textit{half-val} splits as in~\cite{mmtrack2020}

\textbf{Scenario 2 -- from sparse to dense scene:} MOT17~\cite{MOT16} is used as the source train. The target train is MOT20 \textit{half-train} while MOT20 \textit{half-val} is used as a validation set.

\textbf{Scenario 3 -- from surveillance view to drone view:} MOT17~\cite{MOT16} is used as the source train, and the target domain is VisDrone~\cite{visdrone} for pedestrians only. The VisDrone validation set is used to evaluate.

\textbf{Scenario 4 -- from distinguishable appearance to identical appearance:}
MOT17~\cite{MOT16} is used as the source train, and DanceTrack~\cite{sun2022dancetrack} training is set as the target domain. The DanceTrack~\cite{sun2022dancetrack} validation set is used to evaluate.

\subsection{Implement Details}

\begin{algorithm}[!t]
\caption{The training pipeline of UTOPIA}
\label{alg}
\begin{algorithmic}[1]
    \FOR{$\mathbf{x}^t_{src} \in \mathcal{X}_{src}$ and $\mathbf{x}^t_{tgt} \in \mathcal{X}_{tgt}$} %
        \STATE Draw the corresponding $\mathbb{O}_{src}^{t}$
        \STATE Draw $\mathbf{aug} \in \mathbb{Aug}$ and $\mathbf{aug}^{\prime} \in \mathbb{Aug}$
        \STATE Calculate $\ell_{det}^{src} \gets \mathcal{L}_{det}(\mathcal{D}(\mathbf{aug}(\mathbf{x}^t_{src}))) + \mathcal{L}_{det}(\mathcal{D}(\mathbf{aug}^\prime(\mathbf{x}^t_{src})))$
        \STATE Calculate $\ell_{arg}^{src} \gets\mathcal{L}_{agr} = \underset{i}{\texttt{avg}}(1 - \underset{j}{\texttt{max}}(\mathbf{agr}(\mathbf{x}^t_{src})))$
        \STATE Obtain $\mathbb{D}_{src}^{t} \gets \underset{j}{\texttt{argmax}}(\mathbf{agr}(\mathbf{x}^t_{src}))$
        \STATE Sample $\{\mathbf{o}^+_{j}\}, \{\mathbf{o}^-_{j}\} \in \texttt{sampling}(\mathbb{O}_{src}^{t})$
        \STATE Construct the cost matrix $\mathbf{C} (\small{\texttt{sampling}}(\mathbb{O}_{src}^{t}), \mathbb{D}^t_{src})$
        \STATE Obtain the optimal plan $\bar{\pi}$
        \STATE Calculate $\ell_{aux}^{src} \gets \underset{\text{\scriptsize 1:M}}{\mathcal{L}_{aux}} = \bar{\pi}[i, j] - c$ via  Eqn. (18)
        \STATE Calculate $\ell_{MP}^{src} \gets \mathcal{L}_{MP}$ via  Eqn. (17)
        \STATE Optimize $\mathcal{L}_{src} = \ell_{det}^{src} + \ell_{arg}^{src} + \ell_{MP}^{src} + \ell_{aux}^{src}$ \\ w.r.t $\mathbf{x}^t_{src}$
        \STATE Obtain $\widetilde{\mathbb{D}}_{tgt} \gets \underset{j}{\texttt{argmax}}(\mathbf{agr}(\mathbf{x}^t_{tgt}))$
        \STATE Construct the cost matrix $\mathbf{C} (\widetilde{\mathbb{T}}^{t-1}_{tgt}, \widetilde{\mathbb{D}}^t_{tgt})$
        \STATE Obtain the optimal plan $\bar{\pi}$
        \STATE Calculate $\ell_{det}^{tgt} \gets \underset{\bar{\pi}}{\mathcal{L}_{det}}$ via  Eqn. (4)
        \STATE Calculate $\ell_{sim}^{tgt} \gets \underset{\bar{\pi}}{\mathcal{L}_{sim}}$ via  Eqn. (5)
        \STATE Optimize $\mathcal{L}_{tgt} = \ell_{sim}^{tgt} + \ell_{det}^{tgt}$ w.r.t $\mathbf{x}^t_{tgt}$
    \ENDFOR
\end{algorithmic}
\end{algorithm}

\begin{table}[!t]
\centering
\caption{Comparison on augmentation set choices}
\footnotesize
\begin{tabular}{|c|c|c|c|c|}
\hline
& \textbf{MOTA} $\uparrow$ & \textbf{mAP} $\uparrow$ & \textbf{MOTA} $\uparrow$ & \textbf{mAP} $\uparrow$ \\ \hline
\textbf{SET} & \multicolumn{2}{c|}{\textit{MOTSynth $\to$ MOT17}} & \multicolumn{2}{c|}{\textit{MOT17 $\to$ DanceTrack}} \\ \hline
All & 59.40 & 0.673 & 74.3 & 0.778 \\ \hline
Best & \textbf{61.70} & \textbf{0.774} & \textbf{79.6} & \textbf{0.815} \\ \hline
\textbf{SET} & \multicolumn{2}{c|}{\textit{MOT17 $\to$ MOT20}} & \multicolumn{2}{c|}{\textit{MOT17 $\to$ VisDrone}} \\ \hline
All & 55.20 & 0.645 & 13.4 & 0.489 \\ \hline
Best & \textbf{63.90} & \textbf{0.785} & \textbf{16.4} & \textbf{0.651} \\ \hline
\end{tabular}
\label{tab:augmentations}
\end{table}

\begin{figure}[t]
    \centering
    \subcaptionbox{\textit{MOTSynth $\to$ MOT17} \label{fig:recover_example_1}}[0.45\textwidth]{\includegraphics[width=0.45\textwidth]{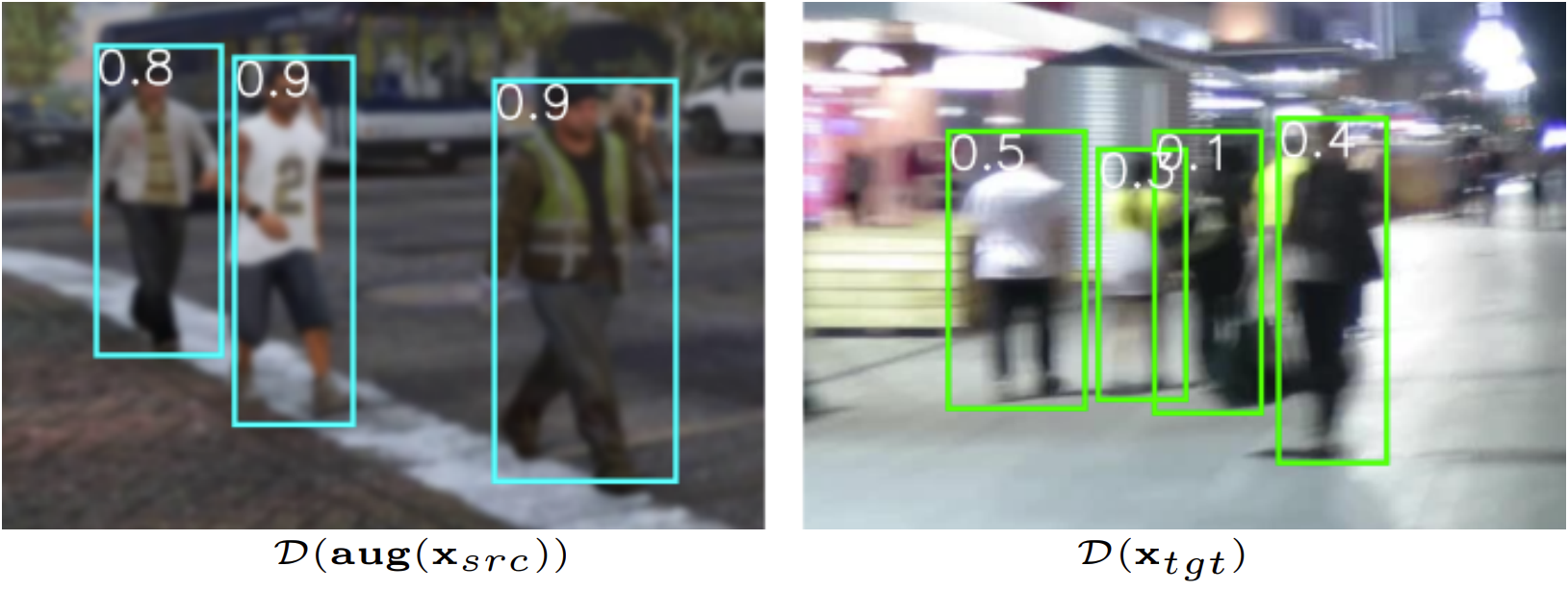}}
    \subcaptionbox{\textit{MOT17 $\to$ VisDrone} \label{fig:recover_example_2}}[0.45\textwidth]{\includegraphics[width=0.45\textwidth]{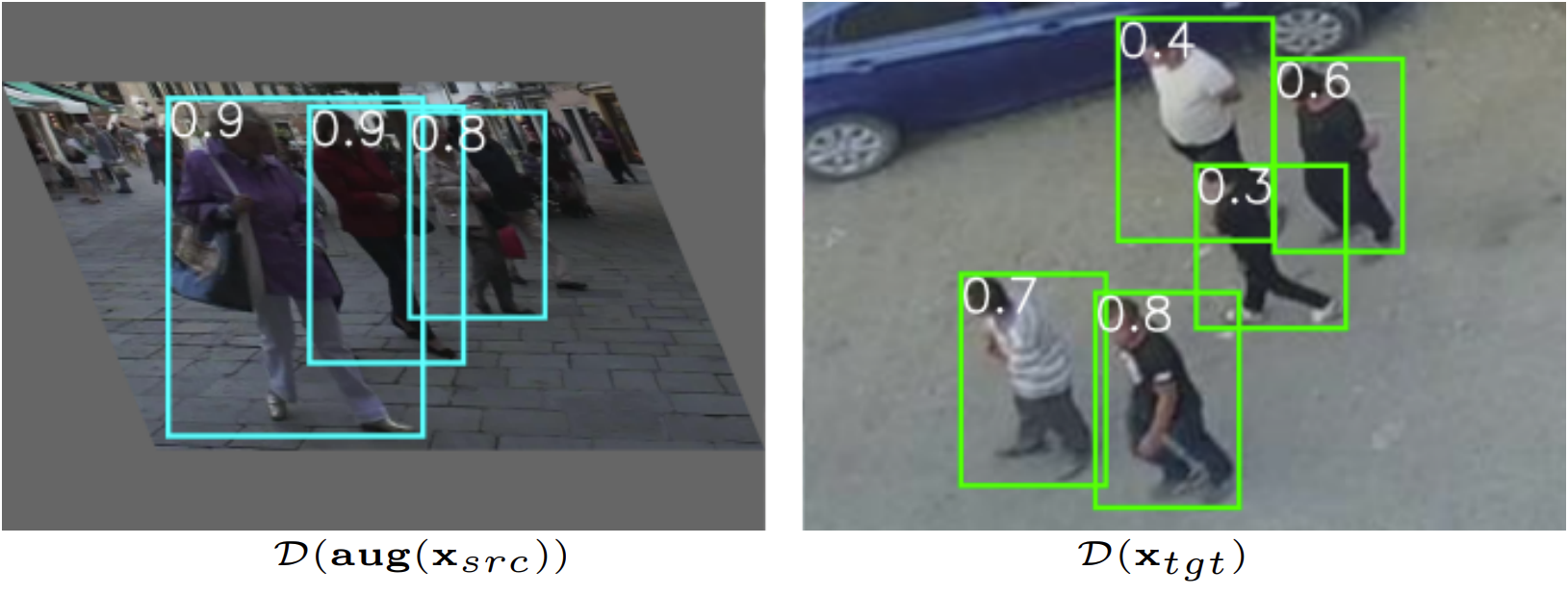}}
    \caption{Adaptively refining the object detector by our proposed agreement can recover low-confident objects. \textbf{Best viewed in color.}}\label{fig:recover_example}
\end{figure}

\begin{figure*}
  \centering
  \subfloat{
    \begin{tikzpicture}
      \begin{axis}[grid=major, width=0.49\textwidth, xlabel=\small{False Positive}, ylabel=\small{False Negative}, label style={font=\small}, tick label style={font=\small}]
        \addplot[smooth, red, mark=*] coordinates
          {(0.290, 0.435) (0.320, 0.284) (0.380, 0.224) (0.484, 0.159)};
        \addplot[smooth, green, mark=*] coordinates
          {(0.185, 0.275) (0.206, 0.175) (0.226, 0.135) (0.256, 0.115)};
        \addplot[only marks, blue, mark=*] coordinates
          {(0.270, 0.354)};
      \end{axis}
    \end{tikzpicture}}%
    \hfill
  \subfloat{
    \begin{tikzpicture}
      \begin{axis}[grid=major, width=0.49\textwidth, xlabel=\small{False Positive}, ylabel=\small{False Negative}, label style={font=\small}, tick label style={font=\small}]
        \addplot[smooth, red, mark=*] coordinates
          {(0.410, 0.475) (0.450, 0.334) (0.530, 0.264) (0.614, 0.249)};
        \addplot[smooth, green, mark=*] coordinates
          {(0.290, 0.315) (0.305, 0.245) (0.326, 0.181) (0.401, 0.164)};
        \addplot[only marks, blue, mark=*] coordinates
          {(0.392, 0.33)};
      \end{axis}
    \end{tikzpicture}} \\
  \subfloat{
    \begin{tikzpicture}
      \begin{axis}[grid=major, width=0.49\textwidth, xlabel=\small{False Positive}, ylabel=\small{False Negative}, label style={font=\small}, tick label style={font=\small}]
        \addplot[smooth, red, mark=*] coordinates
          {(0.450, 0.575) (0.505, 0.483) (0.554, 0.435) (0.674, 0.405)};
        \addplot[smooth, green, mark=*] coordinates
          {(0.290, 0.415) (0.305, 0.373) (0.354, 0.310) (0.475, 0.290)};
        \addplot[only marks, blue, mark=*] coordinates
          {(0.49, 0.437)};
      \end{axis}
    \end{tikzpicture}}
    \hfill
  \subfloat{
    \begin{tikzpicture}
      \begin{axis}[grid=major, width=0.49\textwidth, xlabel=\small{False Positive}, ylabel=\small{False Negative}, legend style={fill=white, fill opacity=0.6, draw opacity=1},text opacity=1, label style={font=\small}, tick label style={font=\small}]
        \addplot[smooth, red, mark=*] coordinates
          {(0.210, 0.365) (0.250, 0.234) (0.310, 0.154) (0.414, 0.089)};
        \addplot[smooth, green, mark=*] coordinates
          {(0.139, 0.275) (0.155, 0.163) (0.175, 0.083) (0.214, 0.049)};
        \addplot[only marks, blue, mark=*] coordinates
          {(0.195, 0.153)};
        \addlegendentry{\small{w/o $\mathcal{L}_{agr}$}}
        \addlegendentry{\small{w/ $\mathcal{L}_{agr}$}}
        \addlegendentry{\small{w/ $\mathcal{L}_{ent}$}}
      \end{axis}
    \end{tikzpicture}}
  \caption{False positive/False negative tradeoff rate measured on four settings: (a) \textit{MOTSynth $\to$ MOT17}, (b) \textit{MOT17 $\to$ MOT20}, (c) \textit{MOT17 $\to$ VisDrone}, (d) \textit{MOT17 $\to$ DanceTrack}. \textbf{Best viewed in color.}}\label{tradeoff}
\end{figure*}
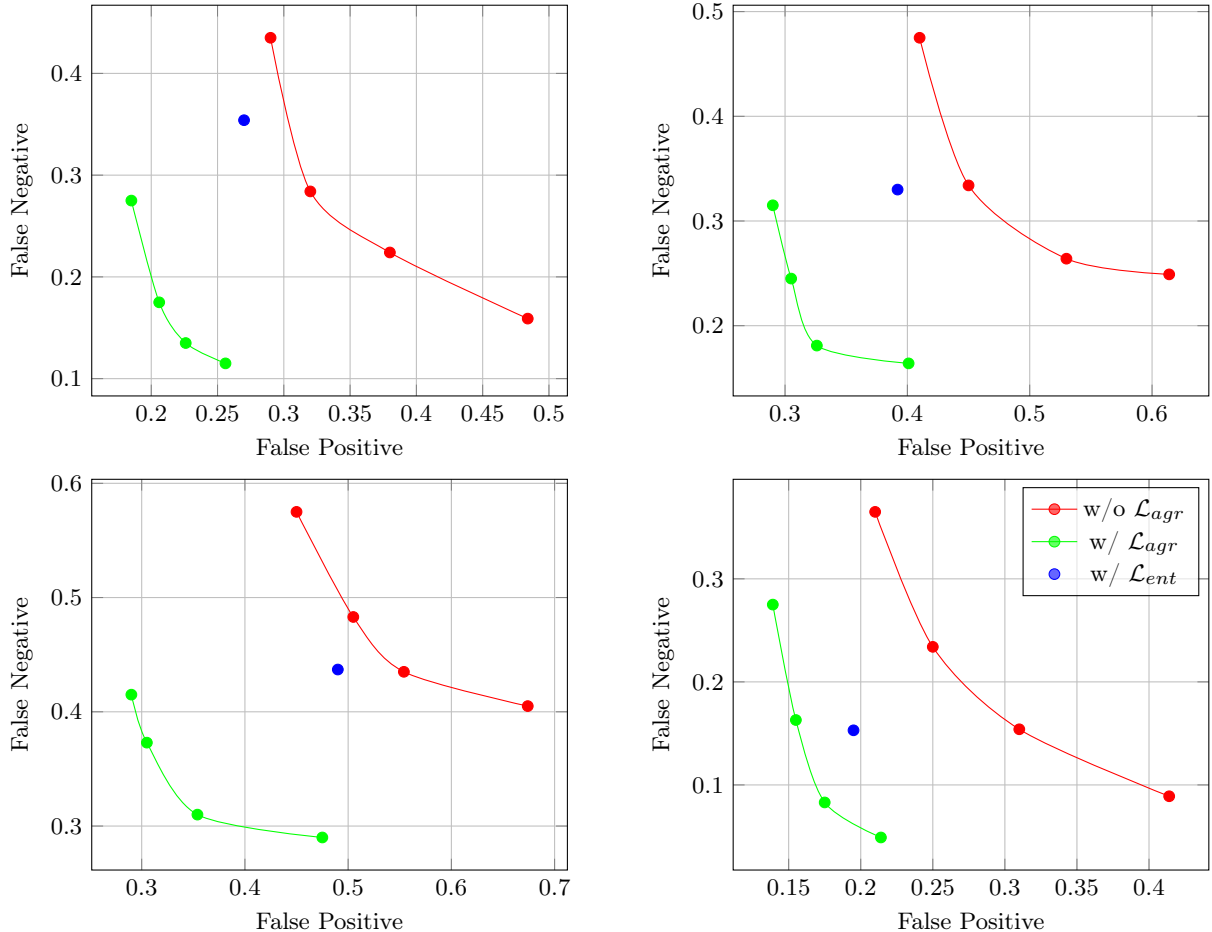

Algorithm~\ref{alg} presents the training process of our proposed framework. We use the mmdetection \cite{chen2019mmdetection} as the base framework, we use
IoU-balanced sampling in that framework to sample RoIs. ResNet-50 \cite{he2015deep} is used as the backbone, and Faster-RCNN \cite{ren2015faster} as the detector. The channel number of embedding features is set to 512. We train our models simultaneously between source and target samples with an initial learning rate of 0.01 for 48 epochs.
To obtain $\widetilde{\mathbb{D}}_{tgt}$, the detection threshold $\widetilde{\gamma} = 0.3$ is the best chosen. Additionally, we used the bi-directional softmax in~\cite{pang2021quasi} as the object-association metric. The track management is the same as the implementation in~\cite{pang2021quasi}.

\subsection{Ablation Study}
\label{subsec:ablation}

\textbf{Augmentation selection} We provide our analyses in Table~\ref{tab:augmentations} to understand the effects of selecting augmentation methods in Eqn.~\ref{eq:costmatrix}. Specifically, we achieve the \textit{Best} accuracies when employed methods implicitly reflect the characteristic transition of the target domain. Particularly, in the \textit{MOTSynth $\to$ MOT17} setting, since the MOT17 has motion blur in moving subjects while objects in MOTSynth are apparent, we simulate the effect by adding random-$\sigma$ Gaussian blur as shown in Fig.~\ref{fig:recover_example_1}. Similarly, we use CutMix~\cite{yun2019cutmix} + color distortion for highly occluded objects in both \textit{MOT17 $\to$ DanceTrack} and \textit{MOT17 $\to$ MOT20} settings. We apply random affine transformations for the \textit{MOT17 $\to$ VisDrone} setting as shown in Fig.~\ref{fig:recover_example_2}, and it requires to do the inverted transforming to the original coordinates before calculating the agreement. \textit{All} means the augmentation composition of color distortion, CutMix~\cite{yun2019cutmix}, Gaussian noise and random affine transformations.

\textbf{False positive / False negative tradeoff}
The detection threshold $\widetilde{\gamma}$ is a sensitive hyper-parameter since it determines the False Negative / False Positive tradeoff rate. To prove the effectiveness of the Object Consistency Agreement strategy, we train the base Faster-RCNN detector~\cite{ren2015faster} and change the threshold from 0.1 to 0.4 to analyze the tradeoff rate, compared to the same detector adding the consistency training. The results are shown in Fig.~\ref{tradeoff}, proving the robustness of the self-trained detector in unseen domains. It is because the detector adaptively learns to recover objects whose scores are lower than $\gamma$. The effect is numerically described in Table~\ref{tab:configs} on MOTA and mAP metrics. We also compare with entropy minimization $\mathcal{L}_{ent}$~\cite{vu2019advent} employed as a soft-label strategy. Since the $\mathcal{L}_{ent}$'s objective is to maximize prediction certainty in the target domain, or other words, it pushes the score to either 0 or 1, ranging scores do not affect the results much, so we choose $\widetilde{\gamma}=0.5$ for $\mathcal{L}_{ent}$ experiments.

\begin{table*}[!t]
 \centering
 \caption{Comparison of configurations. \textbf{Det.} and \textbf{Assg.} columns are experiments for the detection and ID assignment steps, respectively. \xmark \; is the strategy in which the network is only trained on the source domain, while the network in \textbf{Aug} is trained on the augmented source data. \textbf{OCA} and \textbf{OPA} are our proposed self-supervised methods, and \textbf{Sup} stands for fully supervised uses of the ground truth of the target domain.}
 {\begin{tabular}{|c|c|c|c|c|c|c|c|c|c|c|c|c|c|}
 \hline
 \textbf{Det.} & \textbf{Assg.} & \textbf{MOTA} $\uparrow$ & \textbf{IDF1} $\uparrow$ & \textbf{MT} $\uparrow$ & \textbf{ML} $\downarrow$ & \textbf{IDs} $\downarrow$ & \textbf{mAP} $\uparrow$ \\ \hline \hline
 \multicolumn{8}{|c|}{\textit{MOTSynth $\to$ MOT17}} \\ \hline \hline
 \xmark & \xmark & 30.20\% & 38.60\% & 76 & 265 & 1378 & 0.582 \\ \hline \hline
 \textbf{Aug} & \xmark & 30.70\% & 39.20\% & 178 & 148 & 1412 & 0.595 \\\hline
 \textbf{OCA} & \xmark & 38.40\% & 48.70\% & 208 & 88 & 996 & 0.735 \\\hline
 \textbf{OCA} & \textbf{OPA} & \textbf{61.70\%} & \textbf{65.60\%} & \textbf{271} & \textbf{94} & \textbf{468} & \textbf{0.774} \\\hline \hline
 \textbf{Sup} & \xmark & 67.00\% & 71.70\% & 247 & 70 & 346 & 0.876 \\ \hline
 \textbf{Sup} & \textbf{OPA} & \textbf{67.90\%} & \textbf{72.10\%} & \textbf{356} & \textbf{74} & \textbf{343} & \textbf{0.878} \\ \hline \hline
 \multicolumn{8}{|c|}{\textit{MOT17 $\to$ MOT20}} \\ \hline \hline
 \xmark & \xmark & 25.70\% & 23.10\% & 105 & 1037 & 18741 & 0.386 \\ \hline \hline
 \textbf{Aug} & \xmark & 28.30\% & 25.10\% & 118 & 846 & 20845 & 0.476 \\ \hline
 \textbf{OCA} & \xmark & 43.50\% & 36.00\% & 393 & 349 & 19464 & 0.602 \\ \hline
 \textbf{OCA} & \textbf{OPA} & \textbf{63.90\%} & \textbf{50.30\%} & \textbf{908} & \textbf{198} & \textbf{7237} & \textbf{0.785}  \\ \hline \hline
 \textbf{Sup} & \xmark & 55.10\% & 39.40\% & 506 & 411 & 29417 & 0.825 \\ \hline
 \textbf{Sup} & \textbf{OPA} & \textbf{73.90\%} & \textbf{67.10\%} & \textbf{1112} & \textbf{155} & \textbf{2503} & \textbf{0.866} \\ \hline \hline
 \multicolumn{8}{|c|}{\textit{MOT17 $\to$ VisDrone}} \\ \hline \hline
 \textbf{Aug} & \xmark & 10.80\% & 22.30\% & 2 & 62 & 12 & 0.343 \\ \hline
 \textbf{OCA} & \xmark & 15.40\% & 25.60\% & 7 & 48 & 8 & 0.525 \\ \hline
 \textbf{OCA} & \textbf{OPA} & \textbf{16.40\%} & \textbf{26.2\%} & \textbf{9} & \textbf{35} & \textbf{6} & \textbf{0.651} \\ \hline \hline
 \textbf{Sup} & \xmark & 22.70\% & 37.00\% & 11 & 26 & 4 & 0.838 \\ \hline
 \textbf{Sup} & \textbf{OPA} & \textbf{22.80\%} & \textbf{37.20\%} & \textbf{13} & \textbf{24} & \textbf{0} & \textbf{0.851} \\ \hline \hline
 \multicolumn{8}{|c|}{\textit{MOT17 $\to$ DanceTrack}} \\ \hline \hline
 \xmark & \xmark & 38.70\% & 13.50\% & 37 & 55 & 103212 & 0.599 \\ \hline \hline
 \textbf{Aug} & \xmark & 56.20\% & 19.40\% & 82 & 36 & 99328 & 0.721 \\\hline
 \textbf{OCA} & \xmark & 75.40\% & 23.60\% & 188 & 4 & 13177 & \textbf{0.821} \\\hline
 \textbf{OCA} & \textbf{OPA} & \textbf{79.60\%} & \textbf{38.00\%} & \textbf{199} & \textbf{3} & \textbf{6866} & 0.815 \\ \hline \hline
 \textbf{Sup} & \xmark & 72.70\% & 26.10\% & 143 & 13 & 12172 & 0.864 \\ \hline
 \textbf{Sup} & \textbf{OPA} & \textbf{79.70\%} & \textbf{38.80\%} & \textbf{205} & \textbf{3} & \textbf{5499} & \textbf{0.903} \\ \hline
 \end{tabular}}
 \label{tab:configs}
\end{table*}

\textbf{Configurations} We alternatively add and remove the proposed components into the training process and report results in Table~\ref{tab:configs}. Overall, the self-supervised operations \textbf{OCA} and \textbf{OPA} improve the performance of the base strategy \xmark \; on both \textbf{Det.} and \textbf{Assg.} steps. Compared with training on augmented source data only (i.e., \textbf{Aug}), our \textbf{OCA} also takes the target domain feedback into account, resulting in obtaining performance gain over all the settings. On \textit{MOT17 $\to$ MOT20}, \textbf{OCA} gains a $17.8\%$ MOTA increase on \textbf{Det.}, and \textbf{OPA} gains a $14.3\%$ IDF1 increase on \textbf{Assg.}, compared to the \xmark \; one, showing the adaptability on the target domain. On \textit{MOT17 $\to$ DanceTrack}, although it has been proved that an off-the-shelf feature extractor is not always reliable~\cite{sun2022dancetrack}, our adaptable framework can learn to embed discriminative features in pose and shape, result in an enhancement in IDF1 by $12.7\%$, from 26.10\% to 38.80\%. 

\subsection{Comparisons to the State-of-the-Art Methods}
\label{subsec:sota}

\begin{table*}[!t]
 \centering
 \caption{Comparison against State-of-the-arts under the cross-domain setting}
 {\begin{tabular}{|c|c|c|c|c|c|c|c|c|c|c|c|}
 \hline
 \textbf{Type} & \textbf{Method} & \textbf{MOTA} $\uparrow$ & \textbf{IDF1} $\uparrow$ & \textbf{MT} $\uparrow$ & \textbf{ML} $\downarrow$ & \textbf{IDs} $\downarrow$ \\ \hline \hline %
 \multicolumn{7}{|c|}{\textit{MOTSynth $\to$ MOT17}} \\ \hline \hline
 \textbf{Sup} & Trackformer~\cite{meinhardt2022trackformer} & 39.10\% & 51.40\% & 225 & 37 & 870 \\ \hline %
 \textbf{Uns} & ByteTrack~\cite{zhang2021bytetrack} & 41.90\% & 61.0\% & \textbf{336} & \textbf{33} & 797 \\ \hline %
 \textbf{Self} & \textbf{UTOPIA} & \textbf{61.70\%} & \textbf{65.60\%} & 271 & 94 & \textbf{468} \\ \hline \hline %
 \textbf{Self} & Visual-Spatial~\cite{bastani2021self} & 62.10\% & 64.10\% & 229 & 80 & 383 \\\hline
 \textbf{Self} & \textbf{UTOPIA} & \textbf{67.90\%} & \textbf{72.10\%} & \textbf{356} & \textbf{74} & \textbf{343} \\ \hline \hline
 \multicolumn{7}{|c|}{\textit{MOT17 $\to$ MOT20}} \\ \hline \hline
 \textbf{Sup} & Trackformer~\cite{meinhardt2022trackformer} & 36.30\% & 31.30\% & 202 & 686 & 9857 \\ \hline %
 \textbf{Uns} & ByteTrack~\cite{zhang2021bytetrack} & 51.10\% & 49.60\% & 658 & 369 & 4399 \\ \hline %
 \textbf{Self} & \textbf{UTOPIA} & \textbf{63.90\%} & \textbf{50.30\%} & \textbf{908} & \textbf{198} & \textbf{7237} \\ \hline \hline %
 \textbf{Self} & Visual-Spatial~\cite{bastani2021self} & 63.60\% & 64.30\% & 929 & 211 & 2635 \\\hline
 \textbf{Self} & \textbf{UTOPIA} & \textbf{73.90\%} & \textbf{67.10\%} & \textbf{1112} & \textbf{155} & \textbf{2503} \\ \hline \hline
 \multicolumn{7}{|c|}{\textit{MOT17 $\to$ VisDrone}} \\ \hline \hline
 \textbf{Self} & Visual-Spatial~\cite{bastani2021self} &20 .70\% & 32.50\% & 10 & 34 & 5 \\\hline
 \textbf{Self} & \textbf{UTOPIA} & \textbf{22.80\%} & \textbf{37.20\%} & \textbf{13} & \textbf{24} & \textbf{0} \\ \hline \hline
 \multicolumn{7}{|c|}{\textit{MOT17 $\to$ DanceTrack}} \\ \hline \hline
 \textbf{Sup} & Trackformer~\cite{meinhardt2022trackformer} & 69.20\% & 32.30\% & 134 & 8 & 7454 \\ \hline %
 \textbf{Uns} & ByteTrack~\cite{zhang2021bytetrack} & 72.30\% & \textbf{41.20\%} & 176 & \textbf{3} & \textbf{1946} \\ \hline %
 \textbf{Self} & \textbf{UTOPIA} & \textbf{79.60\%} & 38.00\% & \textbf{199} & \textbf{3} & 6866 \\ \hline \hline %
 \textbf{Self} & Visual-Spatial~\cite{bastani2021self} & 73.90\% & 27.90\% & 161 & \textbf{3} & 6357 \\\hline
 \textbf{Self} & \textbf{UTOPIA} & \textbf{79.70\%} & \textbf{38.80\%} & \textbf{205} & \textbf{3} & \textbf{5499} \\ \hline
 \end{tabular}}
 \label{tab:sota}
\end{table*}

\textbf{Cross-domain setting} In Table~\ref{tab:sota}, we compare UTOPIA with different state-of-the-art tracker types: fully-supervised \textbf{Sup}, unsupervised \textbf{Uns} and self-supervised \textbf{Self}. For a fair comparison, in each setting, the first sub-block uses no ground-truth bounding boxes, and the second sub-block is compared with Visual-Spatial~\cite{bastani2021self} using ground-truth bounding boxes. The Visual-Spatial~\cite{bastani2021self} learns an RNN and a Matching Network, it has no self-learning mechanism in object localization, so we have to train with bounding box locations and categorize it into the second sub-block. On \textit{MOT17 $\to$ VisDrone}, only Visual-Spatial~\cite{bastani2021self} is reported since Trackformer~\cite{meinhardt2022trackformer} and ByteTrack~\cite{zhang2021bytetrack} could not perform well without provided ground-truth bounding boxes, returning \texttt{NaN} in most of the metrics. It is worth noting that UTOPIA achieves strong MOTA in most settings. The superior results indicate that UTOPIA is robust to complex cross-scenes. For \textit{MOT17 $\to$ DanceTrack}, UTOPIA shows a lower but comparable IDF1 performance compared with ByteTrack~\cite{zhang2021bytetrack} since diverse non-linear motion patterns in DanceTrack~\cite{sun2022dancetrack} require temporal dynamics to facilitate better association in the tracking process, which we have not addressed it under a self-supervised manner in this work.

\subsection{Qualitative Results}

\begin{figure}
    \centering
    \includegraphics[width=0.45\textwidth]{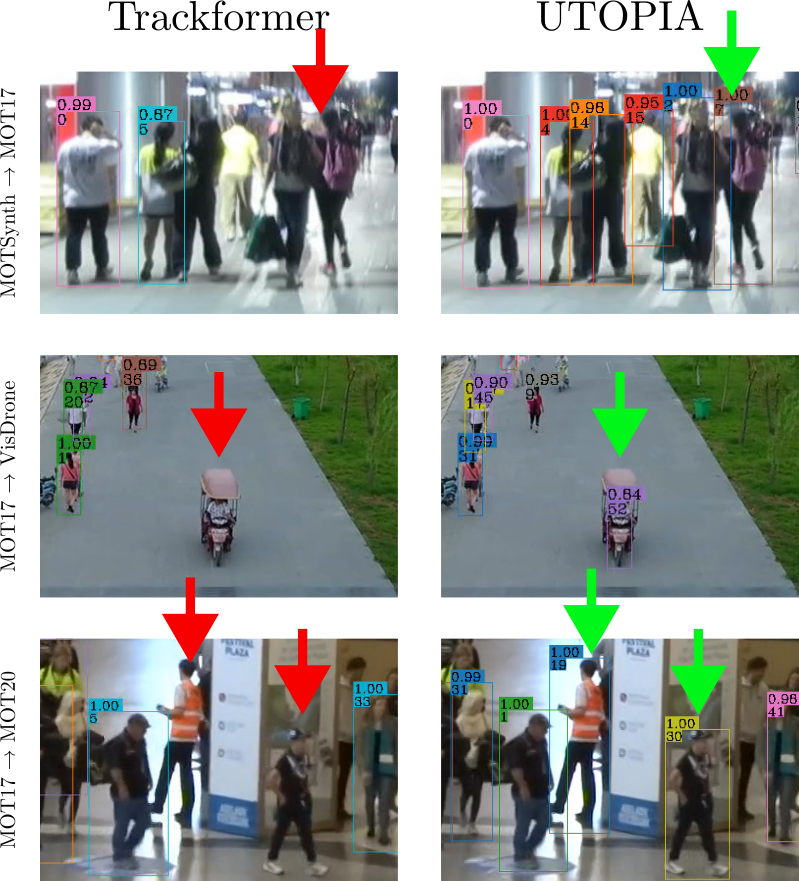}
    \caption{Trackformer \cite{meinhardt2022trackformer} trained on the source domain fails to detect objects, while our UTOPIA can handle these cases. The green arrows indicate the true-positive detection samples; the red arrows indicate the false-negative detection and tracking samples. \textbf{Best viewed in color.}}
    \label{fig:compared}
\end{figure}

\begin{figure}
  \centering
  \subcaptionbox{False-postive cases}[0.45\textwidth]{\includegraphics[width=0.45\textwidth]{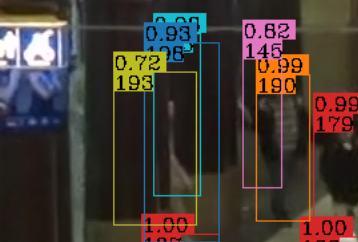}}
  \hfill
  \subcaptionbox{False-negative cases}[0.45\textwidth]{\includegraphics[width=0.45\textwidth]{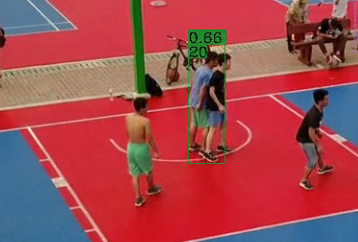}}
  \hfill
  \subcaptionbox{Merging objects error}[0.45\textwidth]{\includegraphics[width=0.45\textwidth]{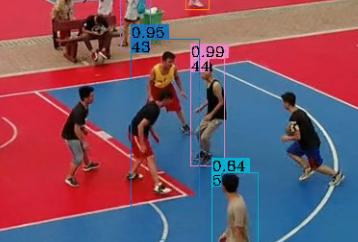}}
  \hfill
\caption{Fail cases. \textbf{Best viewed in color.}}\label{fail}
\end{figure}

Fig. \ref{fig:compared} shows some cases that our UTOPIA can recover from false-negative compared to Trackformer \cite{meinhardt2022trackformer}. 
Fig. \ref{fail} shows some fail cases of our UTOPIA: false-positive, false-negative, and merging objects errors.

\section{Conclusions}
\label{sec:conclusion}
This paper has presented the MOT problem from the cross-domain viewpoint, imitating the process of new data acquisition. Furthermore, it proposed a new MOT domain adaptation without pre-defined human knowledge in understanding and modeling objects. Still, it can learn and update itself from the target data feedback. Through intensive experiments on four \textit{challenging} settings, we first prove the adaptability on self-supervised configurations and then show superior performance on tracking metrics MOTA and IDF1, compared to fully-supervised, unsupervised, and self-supervised methods.

\textbf{Limitations} We acknowledge that the motion model is essential in advanced tracking frameworks. However, this work has not been formulated adaptively in a self-supervised manner, meaning that a motion model could be flexibly integrated. However, it still requires ground truths for fully-supervised training or pre-defined parameters in unsupervised testing.
Moreover, the object type adapted to target data is currently limited to the same object type as source data. The discovery of new kinds of objects is an excellent research avenue for future work.

\section{Data Availability Statement}

The MOT17, MOT20, and MOTSynth datasets analyzed during the current study are available in the \href{https://motchallenge.net/}{MOT Challenge}, an open-access data repository. The dataset includes video and annotations, and it can be accessed at \href{https://motchallenge.net/}{https://motchallenge.net/}. The data is published under the Creative Commons Attribution-NonCommercial-ShareAlike 3.0 License.

The VisDrone dataset analyzed during the current study is available on GitHub. The dataset includes video and annotations, and it can be accessed at \href{https://github.com/VisDrone/VisDrone-Dataset}{https://github.com/VisDrone/VisDrone-Dataset}.

The DanceTrack dataset analyzed during the current study is available on GitHub. The dataset includes video and annotations, and it can be accessed at \href{https://github.com/DanceTrack/DanceTrack}{https://github.com/DanceTrack/DanceTrack}. The data is published under the Creative Commons Attribution-NonCommercial-ShareAlike 4.0 License.

Please note that certain ethical and legal restrictions may apply to the data, and access may require compliance with applicable regulations and obtaining appropriate permissions.

{
    \bibliography{sn-bibliography}
}

\end{document}